\documentclass[conference]{IEEEtran}
\IEEEoverridecommandlockouts
\usepackage{cite}
\usepackage{amsmath,amssymb,amsfonts}
\usepackage{algorithmic}
\usepackage{graphicx}
\usepackage{textcomp}
\usepackage{xcolor}
\def\BibTeX{{\rm B\kern-.05em{\sc i\kern-.025em b}\kern-.08em
		T\kern-.1667em\lower.7ex\hbox{E}\kern-.125emX}}
\begin{document}
	
\title{Enhancing Reasoning Skills in Small Persian Medical Language Models Can Outperform Large-Scale Data Training}
	
\author{\IEEEauthorblockN{1\textsuperscript{st} Mehrdad Ghassabi}
		\IEEEauthorblockA{\textit{Faculty of Computer Engineering} \\
			\textit{University of Isfahan}\\
			Isfahan, Iran \\
			m.ghassabi@eng.ui.ac.ir}
		\and
		\IEEEauthorblockN{2\textsuperscript{nd} Sadra Hakim}
		\IEEEauthorblockA{\textit{School of Computer Science} \\
			\textit{University of Windsor}\\
			Windsor, Canada \\
			hakim6@uwindsor.ca}
		\and
		\IEEEauthorblockN{3\textsuperscript{rd} Hamidreza Baradaran Kashani}
		\IEEEauthorblockA{\textit{Faculty of Computer Engineering} \\
			\textit{University of Isfahan}\\
			Isfahan, Iran \\
			hrb.kashani@eng.ui.ac.ir }
		\and
		\IEEEauthorblockN{4\textsuperscript{th} Pedram Rostami}
		\IEEEauthorblockA{\textit{School of Electrical and Computer Engineering} \\
			\textit{University of Tehran}\\
			Tehran, Iran \\
			pedram.rostami@ut.ac.ir}
                  \and
		\IEEEauthorblockN{5\textsuperscript{th} Zahra Kazemi}
		\IEEEauthorblockA{\textit{Faculty of Computer Engineering} \\
			\textit{University of Isfahan}\\
			Isfahan, Iran \\
			zhrakazemi@mehr.ui.ac.ir }
	}
	
	\maketitle
	
	\begin{abstract}
Enhancing reasoning capabilities in small language models is critical for specialized applications such as medical question answering, particularly in underrepresented languages like Persian. In this study, we employ Reinforcement Learning with AI Feedback (RLAIF) and Direct preference optimization (DPO) to improve the reasoning skills of a general-purpose Persian language model. To achieve this, we translated a multiple-choice medical question-answering dataset into Persian and used RLAIF to generate rejected-preferred answer pairs, which are essential for DPO training. By prompting both teacher and student models to produce Chain-of-Thought (CoT) reasoning responses, we compiled a dataset containing correct and incorrect reasoning trajectories. This dataset, comprising 2 million tokens in preferred answers and 2.5 million tokens in rejected ones, was used to train a baseline model, significantly enhancing its medical reasoning capabilities in Persian. Remarkably, the resulting model outperformed its predecessor, gaokerena-V, which was trained on approximately 57 million tokens, despite leveraging a much smaller dataset. These results highlight the efficiency and effectiveness of reasoning-focused training approaches in developing domain-specific language models with limited data availability.
	\end{abstract}
	
	
	\section{Introduction}

Transformer-based language models
\cite{b1} 
excel at fast, intuitive tasks—such as pattern matching, retrieval, and surface-level text generation—mirroring Kahneman’s concept of “fast thinking”
\cite{b2}.  
However, they struggle with deliberate, multi-step reasoning tasks that require “slow thinking,” particularly in specialized domains such as medicine, where diagnostic accuracy depends on logical inference, evidence evaluation, and error correction.  
This limitation is even more severe in low-resource languages such as Persian, where both high-quality data and compute are scarce.

In 2019, Yoshua Bengio warned that deep learning systems lack true reasoning capacity and called for architectures that support out-of-distribution generalization
\cite{b3}.  
While the transformer architecture was revolutionary, its success has largely come from scaling: larger models and larger datasets yield better performance.  
Yet, despite current large language models having trillions of parameters and being trained on tens of trillions of tokens, they still make surprisingly simple reasoning errors.  
They may even produce inconsistent answers when asked the same question directly or via chain-of-thought (CoT) prompting
\cite{b4}.  
This deficiency becomes even more pronounced in small medical Persian language models, which have far fewer parameters and much less training data.

Recent advances have attempted to improve the reasoning ability of language models—enhancing their performance on “slow thinking” tasks—but these methods often rely on large, well-curated datasets available only for high-resource languages such as English or Chinese.  
In contrast, Persian lacks sufficient high-quality medical datasets.  
To address this gap, we propose a new framework to improve the reasoning capabilities of Persian medical language models under limited data availability.

In our proposed method, we machine-translated an English multiple-choice medical question answering dataset into Persian and applied reinforcement learning from AI feedback (RLAIF)
\cite{b5} and direct preference optimization (DPO)
\cite{b6} to enhance the reasoning ability of a baseline Persian medical model.  
The resulting model is named gaokerena-R.
\footnote{All of our work is open-source and available at github.com/Mehrdadghassabi/Gaokerena-R}

Our prior work, gaokerena-V
\cite{b7}, fine-tuned a Persian medical language model using approximately 57 million tokens (including a portion of a medical corpus and a dataset) via supervised fine-tuning (SFT).  
Although gaokerena-V demonstrated strong medical knowledge, gaokerena-R outperforms it when given chain-of-thought prompts—despite being trained on less medical data.  
Since both models share the same baseline model, aya-expanse-8b
\cite{b8}, we hypothesize that enhancing reasoning skills is more beneficial than scaling data for small, low-resource medical language models.

In summary, our contributions are as follows:

\begin{enumerate}
  \item An efficient RLAIF+DPO framework that generates high-signal CoT preference pairs using a teacher–student loop.
  \item gaokerena-R
\footnote{Available at huggingface.co/gaokerena/gaokerena-r1.0}
, an 8b-parameter medical model demonstrating that reasoning-focused training outperforms data scaling in low-resource medical NLP.
  \item A machine-translated Persian medical multiple-choice question answering dataset designed for reasoning-focused model training.
\end{enumerate}

	\section{Related Work}
To the best of our knowledge, no prior work has focused on developing Persian medical reasoning language models. Existing Persian medical language models, including gaokerena-V, mainly address knowledge representation and general language understanding with limited attention to reasoning.

 L. Pan et al. reviewed several valuable studies on enhancing reasoning capabilities in language models
\cite{b9}.
Accordingly, we focus here on English-language works that offer methodological insights into generating and integrating reasoning data for improving medical language models.
	\subsection{Related Work In Medical Domain}

The MedSSS framework is a notable, self-evolving system designed to instill robust, long-chain reasoning capabilities into small, deployable medical language models, such as the Llama3.1-8B-Instruct base model \cite{b10}. Its core methodology involves leveraging Monte Carlo Tree Search (MCTS) \cite{b11} over diverse medical datasets to construct rule-verifiable reasoning trajectories. These generated paths are used for policy refinement via supervised fine-tuning (SFT) and direct preference optimization (DPO), and to train a unique Process Reward Model (PRM). This PRM employs a soft dual-sided labeling objective that provides crucial step-level supervision, penalizing reasoning steps that degrade node value to mitigate hallucination and enhance interpretability in complex clinical reasoning. The resulting MedSSS system demonstrated state-of-the-art performance across eleven clinical reasoning benchmarks, achieving an average performance gain of +14.12 compared to its base model and significantly surpassing 32B-scale general-purpose reasoning models.


The MedReason \cite{b12} framework is another significant contribution that addresses the scarcity of high-quality medical reasoning data by leveraging a structured medical knowledge graph (KG). This approach converts conventional question–answer (Q\&A) pairs into a dataset of 32,682 factually grounded reasoning trajectories, serving as structural supervision. Fine-tuning models on these paths consistently boosts performance: the framework achieved average accuracy gains of +5.4\% for LLaMA 3.1-Instruct-8B and up to 7.7\% for DeepSeek-Distill-8B. The resulting MedReason-8B model established a new state-of-the-art among 8B-parameter models, notably surpassing the Huatuo-o1-RL-8B by up to 4.2\% on the MedBullets clinical benchmark, validating the importance of KG-based reasoning for interpretability and analytical depth.


The HuatuoGPT-o1 \cite{b13} framework is a two-stage, verification-guided system designed to instill complex reasoning in medical LLMs using 40K verifiable medical problems. Initially, a verifier guides search strategies to generate verified reasoning trajectories for Supervised Fine-Tuning (SFT), teaching the model to refine its answers. Subsequently, Reinforcement Learning (RL) is applied using verifier-based rewards for further enhancement. This methodology resulted in an average 8.5-point performance gain for the 8B model on medical benchmarks, with the complex Chain-of-Thought (CoT) strategy alone providing an average 4.3-point boost.


         \subsection{Related Work In Other Domain}

The DeepSeek-R1 model \cite{b14} demonstrated that applying a reinforcement learning framework, specifically the resource-efficient Group Relative Policy Optimization (GRPO) \cite{b15} , to the DeepSeek-V3-Base model can effectively enhance complex reasoning capabilities. While the pure RL training (DeepSeek-R1-Zero) yielded powerful reasoning behaviors, it simultaneously introduced significant side effects, most notably a deterioration in readability and an increase in language mixing. To counteract these issues, the authors implemented a multi-stage curriculum that incorporated a cold-start supervised fine-tuning (SFT) phase. This targeted intervention proved essential for restoring the model's linguistic quality and coherence. The resultant model, DeepSeek-R1, leveraged this carefully balanced training pipeline to achieve reasoning performance comparable to state-of-the-art models such as OpenAI-o1-1217, thereby confirming that the successful enhancement of reasoning through RL requires a subsequent, targeted fine-tuning stage to maintain linguistic integrity and coherence.


A notable contribution in the area of reasoning enhancement is Thought Preference Optimization (TPO), proposed by Wu et al \cite{b16}. as an iterative Reinforcement Learning from AI Feedback (RLAIF) approach. This preference-based framework compels the model to generate multiple candidate thought and response pairs, subsequently utilizing a judge model to evaluate only the resulting response quality. The collected best–worst pairs are then used to train the baseline model via Direct Preference Optimization (DPO), teaching it to generate high-quality internal thoughts without direct thought supervision. Utilizing an 8B parameter model as the base, this method demonstrated substantial performance gains in general instruction following, achieving an impressive 52.5\% win rate on AlpacaEval (LC) and 37.3\% on Arena-Hard. These results represent gains of over 4\% compared to the direct response baseline, confirming that structured preference optimization effectively enhances reasoning capabilities, even across non-traditional domains such as marketing and health.


The study by N. Ho et al. \cite{b17} introduced Fine-tune-CoT, an effective knowledge distillation method that leverages prohibitively large language models (LLMs) to generate Chain-of-Thought (CoT) rationales for the fine-tuning of significantly smaller student models. This systematic transfer process enabled student models, which were approximately 25–100x smaller in parameter count, to acquire substantial reasoning capabilities. For instance, the 6.7B student model achieved an accuracy of 53.33\% on MultiArith using diverse reasoning, demonstrating that complex, high-performance reasoning abilities can be efficiently transferred across a significant reduction in model scale.


          \section{Proposed Methods}
          
We propose a two-stage framework to enhance the reasoning capabilities of small Persian medical language models. First, we employ Reinforcement Learning with AI Feedback (RLAIF) to construct a preference dataset $\mathcal{D} = \{(x, y_w, y_l)\}$, where $x$ is a medical question, $y_w$ is a preferred reasoning trajectory, and $y_l$ is a rejected one. Second, we fine-tune the student model $\pi_\theta$ using Direct Preference Optimization (DPO).

\begin{figure}[h]
    \centering
    \includegraphics[width=1.0\linewidth]{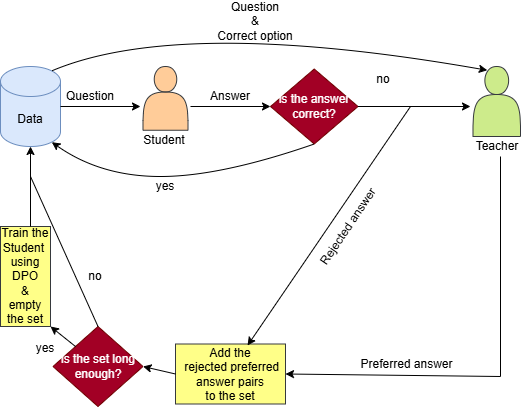}
    \caption{Method 1 Block Diagram}
    \label{fig1}
\end{figure}

\begin{figure}[h]
    \centering
    \includegraphics[width=1.0\linewidth]{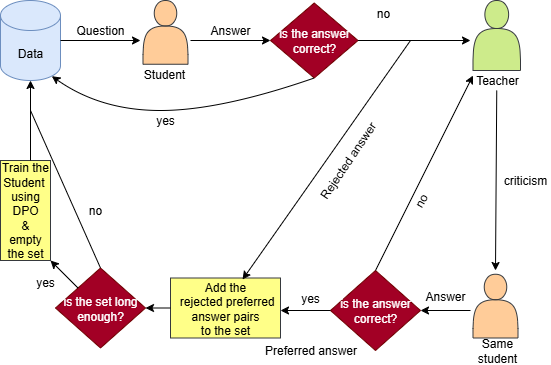}
    \caption{Method 2 Block Diagram}
    \label{fig2}
\end{figure}

To generate high-quality preference pairs, we utilize a teacher-student architecture. The student ($\pi_S$) is the baseline Persian model, and the teacher ($\pi_T$) is a reasoning-capable model (DeepSeek-R). We employ two strategies to populate $\mathcal{D}$ based on the student's performance:
\begin{enumerate}
	\item Teacher Correction for Reasoning Alignment:
	This strategy accounts for 95\% of our training data. The student model $\pi_S$ generates an initial response $y_{init}$ for a given question $x$. If the response is incorrect, $y_{init}$ is designated as the rejected response ($y_l$). The teacher model $\pi_T$ is then prompted with the ground truth answer and instructed to generate a correct, detailed Chain-of-Thought (CoT) explanation. This teacher-generated trajectory serves as the preferred response ($y_w$). The resulting preference pair $(x, y_w, y_l)$ explicitly contrasts flawed student reasoning with expert teacher reasoning. This approach is depicted in Figure \ref{fig1}.
	
	\item Teacher-Guided Self-Correction:
	For the remaining 5\% of the data, we employ an iterative feedback loop to encourage intrinsic error correction. This limited scale was primarily due to hardware constraints. If the student's initial response $y_{init}$ is incorrect, the teacher generates a textual critique $c$ that identifies the error without revealing the correct option. The student is then prompted with $c$ to self-correct and generate a new response, $y_{retry}$. If $y_{retry}$ is correct, it is designated as the preferred response ($y_w$), and the original incorrect response $y_{init}$ remains the rejected response ($y_l$). This method ensures that the final $y_w$ is generated by the student model's own distribution. The full process is detailed in Figure \ref{fig2}.
\end{enumerate}

We optimize the student policy $\pi_{\theta}$ to satisfy the generated preferences using Direct Preference Optimization (DPO). DPO is an alignment method that avoids the need for a separate, expensive reward model. Instead, it optimizes the policy directly by minimizing the negative log-likelihood of the preferred response relative to the reference model $\pi_{ref}$ (the frozen aya-expanse-8b). The objective function is defined as:


\begin{equation*}
	\begin{split}
		\mathcal{L}_{DPO}(\pi_{\theta}; \pi_{ref}) = & -\mathbb{E}_{(x, y_w, y_l) \sim \mathcal{D}} \\
		& \hspace*{-2em} \Biggl[ \log \sigma \Biggl( \beta \log \frac{\pi_{\theta}(y_w|x)}{\pi_{ref}(y_w|x)} - \beta \log \frac{\pi_{\theta}(y_l|x)}{\pi_{ref}(y_l|x)} \Biggr) \Biggr]
	\end{split}
\end{equation*}

Where $\sigma$ is the sigmoid function and $\beta$ is a hyperparameter that controls the extent of the policy's deviation from the reference model. This single objective implicitly maximizes the probability of the valid reasoning trajectory $y_w$ while suppressing the probability of the flawed reasoning path $y_l$. The final dataset comprised approximately 11,000 pairs, yielding about 2 million preferred tokens and 2.5 million rejected tokens.

         \section{Data}
We required a Persian medical multiple-choice question answering dataset to implement our proposed method.  
In the absence of such a dataset at the time, we used DeepSeek-V3
\cite{b18}, a cost-effective large language model, to translate a subset of the MedMCQA dataset from English to Persian.  
To maintain topic diversity, the questions were randomly selected from MedMCQA
\cite{b19}.  
To verify the quality of the translations, we prompted
\footnote{Prompts are available at github.com/Mehrdadghassabi/Gaokerena-R/blob/main/dataset/judgement.ipynb}
 two referees—grok-3-mini
\cite{b20} and gpt-4.1-mini
\cite{b21}—to evaluate each translation.  
A translation was considered verified only if both referees assigned it a score of 5 out of 5.  
This process resulted in approximately 18,000 verified Persian medical multiple-choice questions.

          \section{Carbon Footprint}
\begin{figure}[h]
    \centering
    \includegraphics[width=0.8\linewidth]{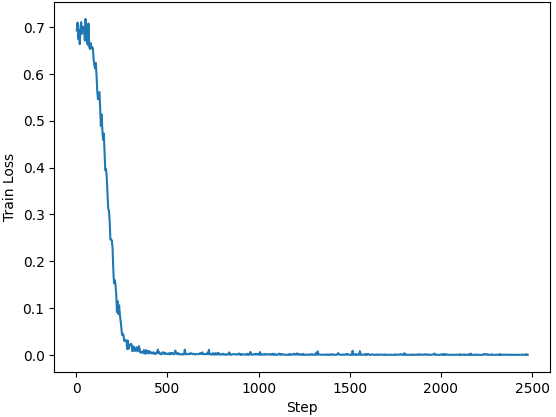}
    \caption{Training loss curve}
    \label{fig3}
\end{figure}

The carbon footprint of our DPO fine-tuning process was estimated based on the hardware configuration and total runtime. The procedure ran for a combined total of 1 hour 
\footnote{
The reported 1 hour refers only to the training time and does not include the time spent on data generation.
}
on an NVIDIA H100 PCIe 80 GB GPU, with approximately 43 GB of VRAM utilized during training. The training loss curve is shown in Figure \ref{fig3}. Assuming an average power consumption of 350 watts per GPU, the total energy usage was approximately 0.35 kWh. Using the average carbon intensity of the Canadian electricity grid, where our server was located (0.086 kilograms of CO2 equivalent per kWh \cite{b22}), this corresponds to an estimated emission of 0.0301 kilograms of CO2 equivalent generated during the fine-tuning process. Compared to our previous model, gaokerena-V, which emitted 2.66 kilograms of CO2, this represents a substantial reduction in environmental impact.\cite{b23}

          \section{Results}
In this section, we compare the newly developed gaokerena-R model with its predecessor, gaokerena-V. While gaokerena-V was trained on a large medical corpus and demonstrates strong factual knowledge and retrieval capabilities, gaokerena-R was specifically designed to enhance medical reasoning. Owing to its reasoning-focused training pipeline, gaokerena-R was trained on a substantially smaller dataset, resulting in slightly reduced coverage of general medical knowledge. However, this trade-off enabled it to develop deeper reasoning competence, allowing it to perform better on tasks requiring multi-step inference and logical consistency.  

In the final evaluation, we compared the performance of gaokerena-V under direct (straight) prompting with that of gaokerena-R when provided with Chain-of-Thought (CoT) prompts. The results highlight that gaokerena-R, despite its smaller scale and limited training data, achieves superior reasoning performance through structured reasoning guidance, demonstrating the effectiveness of reasoning-centered optimization over pure data scaling.
\footnote{You can see some examples that show the enhancements in the Gaokerena-R model at github.com/Mehrdadghassabi/Gaokerena-R/tree/main/assets/examples}
          \subsection{Medical Reasoning Capabillities}
To assess the medical reasoning capabilities of the models, we prompted them to produce Chain-of-Thought reasoning trajectories with a temperature setting of 1.0.  
Since all models can produce different answers for the same question
\footnote{Evaluation prompts are available at github.com/Mehrdadghassabi/Gaokerena-R/blob/main/evaluations/zeroshot-COT/kopp/gaokerena-r1.0/Untitled2.ipynb},
we evaluated them using two metrics on the FA\_MED\_MMLU
\footnote{Available at huggingface.co/datasets/gaokerena/FA\_MED\_MMLU} 
and IBMSEE (September 2023)
\footnote{Available at huggingface.co/datasets/gaokerena/KOPP}
datasets.

              \subsubsection{Accuracy}
The first metric is accuracy for each question in the datasets. Five independent samples were generated per question, and a majority-voting mechanism was applied: if three or more of the five generations selected the same option, that option was chosen as the final prediction; otherwise, the question was left unanswered to reflect model uncertainty.  
This framework provides a robust estimate of reasoning consistency across multiple trajectories.  
Results are presented in Table
\ref{tab:med_reasoning_capabillities_WNM_comparison} (without negative marking) and Table
\ref{tab:med_reasoning_capabillities_NM_comparison} (with negative marking), where, in the negative marking setting, each incorrectly answered question is assigned a score of -0.33.  
These two scoring schemes allow for a fair comparison of the models’ reasoning accuracy under different evaluation criteria.

	\begin{table}[ht]
		\centering
		\caption{Chain-of-Thought Prompted Performance Without Negative Marking}
		\begin{tabular}{|l|c|c|c|}  
			\hline
			\textbf{} & \textbf{gao} & \textbf{gao} & \textbf{aya-} \\ 
			& \textbf{kerena-R} &  \textbf{kerena-V} & \textbf{expanse-8b} \\
			&   & &(baseline)  \\ \hline
			MMLU- &  &  &  \\ 
			anatomy(fa)  & \textbf{42.22} & 39.25  & 40.74 \\ \hline
			MMLU- &    &  &  \\
			medical-genetics(fa) & \textbf{50.0}  & 41.0  & 45.0 \\ \hline
			MMLU- &  &    &  \\
			college-medicine(fa) & 47.97  & 37.57  &\textbf{48.55}  \\ \hline
			MMLU- &    &  &  \\
			clinical-knowledge(fa)& \textbf{55.84} & 46.79  & 54.71  \\ \hline
			MMLU- &  &  &  \\
			professional-& \textbf{44.85} & 37.13 & 43.75 \\
                        medicine(fa)& &  &  \\ \hline
			MMLU- &  &  &  \\
			college-biology(fa)& \textbf{48.61} & 36.80 & 43.75 \\ \hline
			MMLU(avg) & \textbf{48.76} & 40.40 & 47.10 \\ \hline
			IBMSEE Sept2023 & \textbf{38.69}  & 29.76 & 35.71  \\ \hline
			Number of&  &  &  \\
			parameters & 8b & 8b & 8b \\ \hline
			inference time & $\approx 5 \times 35s$ & $\approx 5 \times 35s$ & $\approx 5 \times 35s$ \\  \hline
		\end{tabular}
		\label{tab:med_reasoning_capabillities_WNM_comparison}
	\end{table}

	\begin{table}[ht]
		\centering
		\caption{Chain-of-Thought Prompted Performance With Negative Marking}
		\begin{tabular}{|l|c|c|c|}  
			\hline
			\textbf{} & \textbf{gao} & \textbf{gao} & \textbf{aya-} \\ 
			& \textbf{kerena-R} &  \textbf{kerena-V} & \textbf{expanse-8b} \\
			&   & &(baseline)  \\ \hline
			MMLU- &  &  &  \\ 
			anatomy(fa)  & \textbf{29.13} & 27.65  & 24.93  \\ \hline
			MMLU- &    &  &  \\
			medical-genetics(fa) & \textbf{40.0}  & 32.0  & 33.0  \\ \hline
			MMLU- &  &    &  \\
			college-medicine(fa) & \textbf{34.68}  & 25.24  & 34.48  \\ \hline
			MMLU- &    &  &  \\
			clinical-knowledge(fa)& \textbf{44.65} & 35.59  & 42.51  \\ \hline
			MMLU- &  &  &  \\
			professional-& \textbf{30.39} & 25.0 & \textbf{30.39}   \\
                        medicine(fa)& &  &  \\ \hline
			MMLU- &  &  &  \\
			college-biology(fa)& \textbf{36.80} & 25.0  & 30.09  \\ \hline
			MMLU(avg) & \textbf{36.14} & 28.57  & 33.55  \\ \hline
			IBMSEE Sept2023 & \textbf{24.60}  & 15.87 & 19.84   \\ \hline
			Number of&  &  &  \\
			parameters & 8b & 8b & 8b \\ \hline
			inference time & $\approx 5 \times 35s$ & $\approx 5 \times 35s$ & $\approx 5 \times 35s$ \\  \hline
		\end{tabular}
		\label{tab:med_reasoning_capabillities_NM_comparison}
	\end{table}
           
           \subsubsection{Pass@K}
The pass@k metric, originally introduced by B. Brown et al.
\cite{b24}, 
provides a robust measure for evaluating language models that generate varying outputs for the same input across multiple samples. It quantifies the probability that a model produces at least one correct answer within \(k\) independent attempts, thereby offering a more comprehensive assessment of reasoning reliability and sample diversity. The formal definition of the metric is given in Formula
\ref{fr:passatk}
, where \(N\) denotes the total number of generated samples and \(C_i\) represents the number of correct samples for problem \(i\).

\begin{equation}
\text{pass@k} = \frac{1}{\text{\# of problems}} \sum_{i=1}^{\text{\# of problems}} \left( 1 -  \frac{\binom{N - C_i}{k}}{\binom{N}{k}} \right)
\label{fr:passatk}
\end{equation}

\begin{figure}[h]
    \centering
    \includegraphics[width=1.0\linewidth]{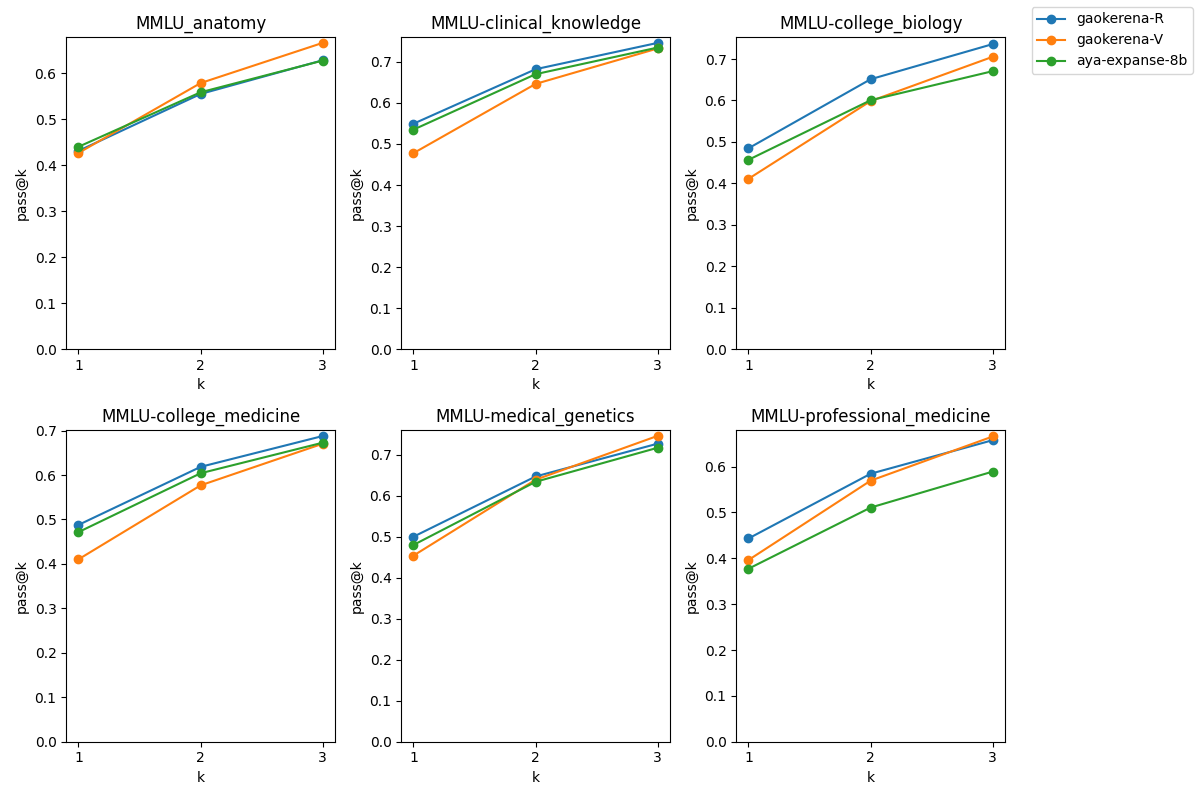}
    \caption{Pass@k results on the FA\_MED\_MMLU dataset using Chain-of-Thought prompting}
    \label{fig4}
\end{figure}

\begin{figure}[h]
    \centering
    \includegraphics[width=0.8\linewidth]{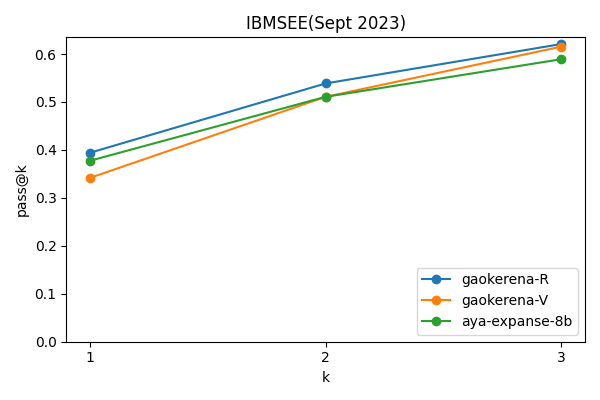}
    \caption{Pass@k results on the IBMSEE (September 2023) dataset using Chain-of-Thought prompting}
    \label{fig5}
\end{figure}

Following the same experimental setup described earlier, we computed pass@k scores for \(k = 1, 2, 3\) using both the FA\_MED\_MMLU dataset and the IBMSEE (September 2023) dataset. The evaluation included the gaokerena-R, gaokerena-V, and aya-expanse-8b models. The results for the FA\_MED\_MMLU dataset are illustrated in Figure
\ref{fig4}, while the IBMSEE (September 2023) results are shown in Figure
\ref{fig5}.

As illustrated in Figures
\ref{fig4} and
\ref{fig5}, the gaokerena-R model consistently demonstrated superior performance across nearly all \(k\) values and evaluation categories. This improvement indicates that gaokerena-R possesses more stable and coherent reasoning capabilities, producing correct answers more reliably even with a limited number of samples. In contrast, gaokerena-V exhibited relatively weak performance for smaller \(k\) values, while its results improved as \(k\) increased. This pattern suggests that gaokerena-V is considerably more uncertain when prompted to generate Chain-of-Thought reasoning trajectories, often producing diverse or inconsistent answers across different samples. The results therefore highlight the enhanced reasoning consistency and reliability achieved through gaokerena-R’s targeted reasoning-oriented training approach.

           \subsection{Medical Knowledge}
As mentioned earlier, gaokerena-V possesses a broader base of medical knowledge compared to the newer gaokerena-R model, primarily because it was trained on a significantly larger and more diverse corpus of medical data.  
This difference becomes evident when both models are evaluated using direct question–answer prompting, rather than being asked to generate explicit chain-of-thought reasoning trajectories.  
As shown in Table
\ref{tab:med_knowledge_comparison}, under this direct prompting setting, gaokerena-V demonstrates superior performance, reflecting its stronger memorization and factual recall abilities derived from large-scale medical pretraining.  
Interestingly, the approximately equal performance of gaokerena-R and its baseline aya-expanse-8b in this same direct prompting setup indicates that gaokerena-R’s superior results in the chain-of-thought setting stem primarily from its enhanced reasoning skills rather than from increased medical knowledge.

	\begin{table}[ht]
		\centering
		\caption{Straight Prompted Performance}
		\begin{tabular}{|l|c|c|c|}  
			\hline
			\textbf{} & \textbf{gao} & \textbf{gao} & \textbf{aya-} \\ 
			& \textbf{kerena-R} &  \textbf{kerena-V} & \textbf{expanse-8b} \\
			&   & &(baseline)  \\ \hline
			MMLU- &  &  &  \\ 
			anatomy(fa)  & 41.48 & \textbf{48.14}  & 40.74  \\ \hline
			MMLU- &    &   &   \\
			medical-genetics(fa) & 49.0  & \textbf{53.0}  &  49.0 \\ \hline
			MMLU- &  &    &  \\
			college-medicine(fa) & \textbf{46.24} & 43.93  & 44.51   \\ \hline
			MMLU- &    &  &  \\
			clinical-knowledge(fa) & 52.45 & \textbf{55.47}  & 52.07  \\ \hline
			MMLU- &  &  &  \\
			professional-& 41.91  & \textbf{47.05}  & 45.58   \\
                        medicine(fa)& &   &   \\ \hline
			MMLU- &  &  &  \\
			college-biology(fa)& 44.44 & \textbf{47.22}  &  45.14 \\ \hline
			MMLU(avg) & 46.28 & \textbf{49.31}  & 46.64 \\ \hline
			IBMSEE Sept2023 & 35.11  &\textbf{38.69} & 34.52  \\ \hline
			Number of&  &  &  \\
			parameters & 8b & 8b & 8b \\ \hline
			inference time & $\approx10s$ & $\approx 10s$ & $\approx 10s$ \\  \hline
		\end{tabular}
		\label{tab:med_knowledge_comparison}
	\end{table}

           \subsection{Final Evaluation}

As previously discussed, gaokerena-V performs better when prompted directly, whereas gaokerena-R excels when prompted with a Chain-of-Thought (CoT) format that encourages step-by-step reasoning before producing an answer. This observation indicates a high degree of prompt dependence, a key characteristic of both models in their current state.

One key advantage of CoT prompting—particularly when combined with multiple sampling and majority voting—is that it provides a natural measure of model certainty. If the generated responses converge on the same option across samples, the model can be considered confident; conversely, a wide dispersion among responses indicates uncertainty. This approach offers a practical alternative to directly querying the model’s self-assessed confidence
\cite{b25}
, a capability that smaller models generally lack due to their limited self-awareness and introspective reasoning abilities.

In cases where gaokerena-R exhibits uncertainty (i.e., when multiple distinct answers are produced), we employ the baseline model, aya-expanse-8b, as an auxiliary verifier. The outputs from gaokerena-R are presented to aya-expanse-8b, which is tasked with selecting the option containing the least incorrect or inconsistent information. This hybrid evaluation framework enables us to combine the reasoning strengths of gaokerena-R with the broader knowledge coverage of aya-expanse-8b. 

Accordingly, we compare two configurations, as summarized in Table
\ref{tab:med_opns_comparison}.
 In the first configuration , gaokerena-V is evaluated using direct prompting without any reasoning guidance. In the second configuration , gaokerena-R is prompted with a Chain-of-Thought (CoT) format, and in cases of uncertainty, its generated answers are verified by aya-expanse-8b to identify the most reliable response.

	\begin{table}[ht]
		\centering
		\caption{Evaluation of two different configurations}
		\begin{tabular}{|l|c|c|}  
			\hline
			\textbf{} & \textbf{gaokerena-R} & \textbf{gaokerena-V}  \\ 
			& \textbf{+} &   \\
			& \textbf{ aya-expanse-8b}  &     \\ 
                        & \textbf{(verifier)}  &     \\ \hline
			MMLU- &  &    \\ 
			anatomy(fa)  & 47.40  & \textbf{48.14}   \\ \hline
			MMLU- &   &      \\
			medical-genetics(fa) & \textbf{56.0}  & 53.0   \\ \hline
			MMLU- &  &      \\
			college-medicine(fa) & \textbf{50.28} & 43.93    \\ \hline
			MMLU- &    &    \\
			clinical-knowledge(fa) & \textbf{58.86}  & 55.47  \\ \hline
			MMLU- &  &    \\
			professional-& \textbf{48.89} & 47.05  \\
                        medicine(fa)& &      \\ \hline
			MMLU- &  &   \\
			college-biology(fa)& \textbf{54.86} & 47.22   \\ \hline
			MMLU(avg) & \textbf{52.98}  & 49.31   \\ \hline
			IBMSEE Sept2023 & \textbf{46.42}  &38.69   \\ \hline
                        prompt & COT for the main model & Straight   \\ 
                        &            Straight for the verifier   &   \\ \hline
			inference time & $\approx 5 \times 35 + 10 + 8 s$ & $\approx 10s$  \\  \hline
		\end{tabular}
		\label{tab:med_opns_comparison}
	\end{table}
        
\section{Future Research}
The results show that gaokerena-V performs better with direct prompts, while gaokerena-R excels with chain-of-thought (CoT) prompts.  
This indicates that both models are prompt-dependent, and future work must address this critical limitation.

Specifically, future research should aim to develop prompt-invariant medical language models that integrate strong reasoning skills and medical knowledge, achieving superior performance regardless of the prompt format. This involves resolving the inherent trade-off between the superior factual recall of a data-scaled model (like gaokerena-V) and the enhanced logical consistency of a reasoning-focused model (like gaokerena-R). Achieving prompt invariance would represent an important step toward more reliable and generalizable small Persian medical language models.


\end{document}